\documentclass[3p]{elsarticle}
\graphicspath{ {./figures/} }
\usepackage{hyperref}
\usepackage{float}
\usepackage{verbatim} 
\usepackage{apalike}
\usepackage{enumitem}
\usepackage{todonotes}

\usepackage{subfigure}
\usepackage{amsmath} 
\restylefloat{figure}
\restylefloat{table}

\newcommand{\revised}[1]{#1}
\newcommand{\rev}[1]{#1}
\journal{}

\biboptions{authoryear}

\begin{document}
\begin{frontmatter}









\title{Calibrated Explanations: \\with Uncertainty Information and Counterfactuals}

\author[label1,label2]{Helena Löfström \corref{cor1}}
\ead{helena.lofstrom@ju.se}

\author[label3]{Tuwe Löfström}
\ead{Tuwe.lofstrom@ju.se}

\author[label3]{Ulf Johansson}
\ead{Ulf.johansson@ju.se}

\author[label3]{Cecilia Sönströd}
\ead{Cecilia.sonstrod@ju.se}

\cortext[cor1]{Corresponding author.}
\address[label1]{Jönköping International Business School, Jönköping University, Sweden}
\address[label2]{Department of Information Technology, University of Borås, Sweden}
\address[label3]{Department of Computing, Jönköping University, Sweden}

\begin{abstract}
While local explanations for AI models can offer insights into individual predictions, such as feature importance, they are plagued by issues like instability. The unreliability of feature weights, often skewed due to poorly calibrated ML models, deepens these challenges. Moreover, the critical aspect of feature importance uncertainty remains mostly unaddressed in Explainable AI (XAI). The novel feature importance explanation method presented in this paper, called Calibrated Explanations (CE), is designed to tackle these issues head-on. Built on the foundation of Venn-Abers, CE not only calibrates the underlying model but also delivers reliable feature importance explanations with an exact definition of the feature weights. CE goes beyond conventional solutions by addressing output uncertainty. It accomplishes this by providing uncertainty quantification for both feature weights and the model's probability estimates. Additionally, CE is model-agnostic, featuring easily comprehensible conditional rules and the ability to generate counterfactual explanations with embedded uncertainty quantification. Results from an evaluation with 25 benchmark datasets underscore the efficacy of CE, making it stand as a fast, reliable, stable, and robust solution.
\end{abstract}


\end{frontmatter}

\section{Introduction} \label{introduction}
\noindent \rev{A core task in many AI-based systems is predictive modelling, in which a machine learning (ML) algorithm is used to create a model, mapping input to target from historical examples to be able to predict novel inputs.} Predictive models used for AI-based decision support are generally not designed for transparency. Although they operate in critical situations such as, e.g., medicine or defence, they are limited to only presenting a probable outcome \citep{DavidGunning2017,Ribeiro2016_kdd}, which can lead to either misuse (based on user reliance being higher than appropriate) or disuse (due to users having less reliance than appropriate) \citep{alvarado2014reliance,buccinca2020proxy}.

Due to the lack of transparency, predictions from this type of model often require an explanation. In \textit{explainable artificial intelligence} (XAI), the goal is to create methods that help human users identify when to trust a prediction and when not to, such as an erroneous prediction in a medical diagnosis \citep{marx2023but}. An \textit{explanation} should reveal the strengths and weaknesses of the underlying model to communicate how they will behave in the future \citep{DavidGunning2017, dimanov2020you}.

There are two main categories of explanations: local explanations, which present information about the reasons for individual predictions, and global explanations, which provide information about the general behaviour of the model \citep{guidotti2018survey,moradi2021post,Martens14}. 
\rev{Local explanations have several drawbacks, such as instability, where slight variations in an instance can lead to significantly different explanations and create issues in evaluating the quality of explanations \citep{slack2021reliable, rahnama2019study}. }

\rev{Post hoc explanations can be either \textit{factual} (the feature value causes this effect on the prediction) or \textit{counterfactual} (what are the minimal changes to an instance's feature values that change the prediction) \citep{mothilal2020explaining, guidotti2022counterfactual, wachter2017counterfactual}. Counterfactual explanations are seen as highly human-friendly since they mirror how humans reason \citep{molnar2020interpretable}.}

Another limitation of today's state-of-the-art explanation methods is that they offer limited insights into model uncertainty and reliability. Although earlier authors have pointed out that uncertainty information is essential for decision-making (see, e.g., \citep{parmigiani2002measuring}), it is only in recent years that uncertainty estimation has been identified as a necessary part of an explanation to make the underlying model more transparent \citep{bhatt2021uncertainty, slack2021reliable,antoran2020getting}. \cite{zhou2021evaluating} point out that the primary problem for users of AI is not primarily the model's opaqueness or complexity but rather the output uncertainty. \cite{phillips2020decision} highlight user uncertainty as one of four stress factors that affect decision quality. Consequently, accurate uncertainty estimation can improve the quality and usefulness of explanations in XAI. \rev{Although uncertainty is highlighted as a critical factor in transparent decision-making, there are high costs of obtaining calibrated uncertainty estimates for complex problems \citep{bhatt2021uncertainty,marx2023but}.}

In local explanation methods, the probability estimate that most ML models output is commonly used as an indicator of the likelihood of each class. However, it is well-known that ML models are often poorly calibrated, meaning that the probability estimates do not correspond to the actual probabilities of being correct \citep{van2019calibration,grushka2017ensembles}. Calibration methods like Venn-Abers (VA) \citep{vovk2012vennabers} have been developed to address these issues. VA produces a probability interval for each prediction, which can be aggregated into a calibrated probability estimate through regularisation to compare with other calibration methods or the underlying model's probability estimate.

When using VA for decision-making, it is crucial to note that the technique also provides intervals for each class, quantifying the uncertainty in the probability estimate, which is valuable from an explanatory perspective. The width of the interval reflects the model's uncertainty, where a narrower interval indicates less uncertainty in the probability estimate and a wider interval indicates more uncertainty in the probability estimates. 

\rev{A core assumption of this paper is that it is of questionable value to use an explanation of a poorly calibrated model to support decision-making, as the poorly calibrated model will not reflect reality. The use of explanations of such models will likely eventually result in disuse since the explanations will be misleading in relation to reality. It is also assumed to be beneficial for understanding an explanation that it takes the form of conditional rules, clarifying under which conditions it applies and making it clearly interpretable. Furthermore, having an exactly defined meaning of the feature weight in relation to the prediction of the model is important. This paper applies to binary classification but focuses on explaining probability estimates rather than class labels, as probability estimates carry richer information about the prediction. Quantifying the uncertainty of the feature weight is an important tool to help users drill into a deeper understanding of the prediction. Finally, being able to explore both factual and counterfactual explanations easily is seen as beneficial from a decision-making perspective.}

\rev{This paper introduces a new local feature importance explanation method called Calibrated Explanations. Calibrated Explanations has the following key characteristics:
\begin{enumerate}
    \setlength{\itemsep}{1pt}
    \setlength{\parskip}{0pt}
    \setlength{\parsep}{0pt}
    \item As a pre-requisite for constructing the explanations, the underlying model is calibrated to ensure alignment with reality. VA provides uncertainty estimation for the calibrated probability estimates, and this information is conveyed to the user.
    \item The Calibrated Explanations method provides feature weights with an exact meaning in relation to the calibrated probability estimate of the model. The feature weights indicate the importance of that feature for explaining the prediction.
    \item The feature weights are accompanied by uncertainty quantification, precisely defined in relation to the underlying model's calibrated probability estimates.
    \item Conditional rules are defined so that they have a straightforward interpretation in relation to instance values.
    \item Calibrated Explanations support both factual and counterfactual explanations based on the model's calibrated probability estimates.
\end{enumerate}}

The paper is organized as follows: The next section reviews fundamental concepts related to explanation methods\rev{, calibration,} and VA predictors. The main contribution is introduced in section~\ref{CE}, \rev{defining} Calibrated Explanations. \rev{Section~\ref{presentation} provides a presentation of both factual and counterfactual explanations. In section~\ref{evaluation}, the setup and results from two experiments are presented}. The paper ends with a discussion followed by concluding remarks.


\section{Background}


\subsection{Desiderata of Explanations} \label{sec:ess_char}
\noindent Creating high-quality explanations in XAI depends on the goals it addresses, which may vary, e.g. assessing how the users appreciate the explanation interface is distinct from evaluating if the explanation accurately mirrors the underlying model \citep{Lofstrom2022}. Nonetheless, specific characteristics are universally desirable for post-hoc explanation methods:
\begin{itemize}
    \setlength{\itemsep}{1pt}
    \setlength{\parskip}{0pt}
    \setlength{\parsep}{0pt}
    \item It is critical that an explanation method \textit{reliably} mirror the underlying model, which is closely connected to the concept that an explanation method should have a high level of \textit{fidelity} to the underlying model \citep{slack2021reliable,murdoch19,Carvalho19,arrieta20,das2020opportunities,chromik2020taxonomy,hoffman2018metrics,doshi2017towards, adadi2018peeking, wang2019Designing}. In that sense, a trustworthy explanation must have feature weights that accurately correspond to the probability estimates, \rev{which in turn must accurately reflect the reality, i.e., the model must} be \textit{well-calibrated} \citep{bhatt2021uncertainty}. 
    \item \textit{Stability} and \textit{robustness} are two additional critical features of explanation methods \citep{dimanov2020you, agarwal2022openxai, alvarez2018robustness, chromik2020taxonomy}. Stability refers to the consistency of the explanations; the same instance and model should produce identical explanations across multiple runs \citep{slack2021reliable, carvalho2019machine}. Stability is pointed out as one of the essential characteristics of any explanation technique \rev{by} \cite{zhou2021s}, and the authors write that an unstable method is considered unreliable \rev{since it} provides little insight into how the model works \rev{\citep{slack2021reliable, moradi2021post, hoffman2018metrics, carvalho2019machine, adadi2018peeking, wang2019Designing, mueller19, agarwal2022openxai}}. In contrast, robustness refers to the ability of an explanation method to produce consistent results even when an instance or \rev{its environment }undergoes small perturbations \citep{dimanov2020you, doshi2017towards}. 
\end{itemize}

Consequently, the essential characteristics of an explanation method in XAI are that it should be reliable, stable, and robust.

Two of the most well-established local feature importance techniques are Local Interpretable Model-agnostic Explanations (LIME) \citep{Ribeiro2016_kdd}, and SHapley Additive exPlanations (SHAP) \citep{lundberg2017unified}. LIME uses a surrogate model, which tries to explain each instance using a weighted linear model locally. It perturbs input data and observes the impact on model predictions to create a locally faithful explanation. Each feature explanation consists of a condition and a weight. The technique is sensitive to perturbation choices, which can cause instability and decreased robustness. The problem with instability in LIME has been highlighted in several studies \citep{rahnama2019study,zhou2021s, agarwal2022openxai}.

SHAP values are based on cooperative game theory and provide a unified way to explain the impact of each feature on a model's prediction by considering all possible feature combinations. As opposed to LIME, SHAP does not provide any rules or conditions. Instead, the explanation for any specific instance is in relation to a baseline, which could create interpretation challenges \citep{molnar2020interpretable}. On the other hand, this also makes it possible to aggregate the explanations to gain global explanations. Calculating SHAP values is computationally intensive, especially for high-dimensional data, and the complexity increases with the number of features. As a consequence, it may require specialized libraries to implement effectively.

While LIME and SHAP have emerged as two of the most used and known explanation methods, there are other explanation methods proposed that address different gaps, such as, e.g., Anchors for local Explanations (ANCHOR) \citep{ribeiro2018anchors}, Model Agnostic suPervised Local Explanations (MAPLE) \citep{plumb2018model}, and LOcal Rule-based Explanations (LORE) \citep{guidotti2018local}. ANCHOR identifies local rules or conditions and focuses on finding the most salient features and their thresholds. MAPLE is a supervised neighbourhood approach that combines ideas from local linear models and ensembles of decision trees. LORE aims to create interpretable, human-readable rules that describe why a particular prediction was made through counterfactuals. \rev{Model-Agnostic Counterfactual Explanations (MACE) \citep{karimi2020model} is another counterfactual method that try to identify the set of features resulting in the desired prediction, while remaining at minimum distance from the original set of features describing the instance. An interesting approach, with a framework based on Naïve Bayes, is presented by \cite{slack2021reliable} and instantiated in SHAP (BayesSHAP) and LIME (BayesLIME) to generate well-calibrated explanations with uncertainty information.}

\subsection{Calibration and Venn-Abers predictors}\label{VennAbers}
\noindent A probabilistic predictor generates a predicted class label and a probability distribution over all possible labels. To determine the validity of the predicted probability distributions, statistical tests are performed based on observation of the actual labels. Achieving validity for probabilistic predictions is generally impossible, as discussed by \cite{vovk2005algorithmic}. However, this paper focuses on calibration, which can be defined as follows:

\begin{equation}
\label{calibration}
p(c \mid p^{c})\approx p^{c}\revised{,}
\end{equation}
where $p^{c}$ represents the probability estimate for a particular class label $c$. A well-calibrated model produces predicted probabilities that match observed accuracy. \rev{This means that} if a model assigns a probability estimate of $0.9$ to a specific label, that label should be correct approximately $90\%$ of the time. A well-calibrated model will also produce well-calibrated probability estimates for each instance and class. However, many predictive models produce poorly calibrated probability estimates \citep{van2019calibration}. When a model is poorly calibrated, an external calibration method can be applied using a separate portion of the labelled data, called the calibration set, to adjust the predicted probabilities.

\textit{Venn predictors} are probabilistic predictors \citep{vovk2004selfcalibrating} that, for each label, output multi-probabilistic predictions. These multi-probabilistic predictions are converted into probability intervals where the interval size indicates confidence in the estimation.

In \textit{inductive Venn prediction} \citep{Lambrou2015}, the underlying model is used to divide the calibration instances into a number of \textit{categories} based on a \textit{Venn taxonomy}. Within each category, the estimated probability for test instances falling into a category is the relative frequency of each class label among all calibration instances in that category. Validity is achieved by including the test instance in the calculation. Since the correct label is unknown for the test instances, every possible label is tried, resulting in the multi-probabilistic probability distribution. 

Choosing an appropriate taxonomy when using Venn predictors can be challenging. One alternative is \textit{Venn-Abers predictors} (VA) \citep{vovk2012vennabers}, where the taxonomy is automatically optimized using \textit{isotonic regression}. Isotonic regression is a non-parametric regression technique that fits a piecewise constant function to the data, such that the function is monotonically increasing or decreasing. VA predictors are used together with \textit{scoring classifiers}, and since VA predictors are Venn predictors, they inherit the validity guarantees.

To define a VA calibrator, let $Z_T=\{z_1, \dots, z_{n}\}$, where $n=l+q$, be a training set. Each instance ${z_i=(x_i,y_i)}$ consists of two parts; an \textit{object} $x_i$ and a \textit{label} $y_i$. Since a calibration set is required for the calibration, the training set is divided into a proper training set $Z_q$ and a calibration set $Z_l=\{z_{1},\dots,z_l\}$. Train a scoring classifier that outputs a \textit{prediction score} $s(x)$ when predicting an object $x$. A higher value of $s(x)$ signals a more pronounced belief in the positive class, i.e., class label $1$. To obtain the predicted class label from a scoring classifier, the score is compared to a fixed threshold value $t$. The prediction gets the value of $1$ if $s(x)>t$, and $0$ otherwise. When using VA with scoring classifiers, the threshold is not a fixed value $t$. Instead, an isotonic regression model is fitted to a number of prediction scores where the true targets are known, which creates an increasing function $g(s(x))$ \citep{zadrozny-elkan01}, interpreted as the probability that the label of $x$ is $1$, i.e., a calibrator. 

Thus, an inductive VA predictor produces a multi-probabilistic prediction for a test object $x$\footnote{For convenience, the index $n+1$ is omitted from the test instance to reduce clutter.} with corresponding unknown label $y$ as follows: 
\begin{enumerate}
    \setlength{\itemsep}{1pt}
    \setlength{\parskip}{0pt}
    \setlength{\parsep}{0pt}
    \item Let $g_0$ and $g_1$ be isotonic calibrators for $\{(s(x_{1}),y_{1}),\dots,(s(x_l),y_l),(s(x),0)\}$ and $\{(s(x_{1}),y_{1}),\dots,(s(x_l),y_l),(s(x),1)\}$, respectively.
    \item Let the probability interval for $y=1$ be $[g_0(s(x)),g_1(s(x))]$ (henceforth referred to as $[p_0,p_1]$).
    \item When a regular probability estimate is needed, the probability intervals $[p_0, p_1]$ are aggregated into a single \textit{calibrated probability estimate} by following the recommendation from \cite{vovk2012vennabers} to get a regularized value: 
        \begin{equation} 
        \label{eq:regularize_VA} 
        p=\frac{p_1}{1-p_0+p_1}
        \end{equation}       
\end{enumerate}

In summary, VA produces a calibrated probability estimate (the regularized $p$-value) and a probability interval $[p_0,p_1]$.

\section{Definition of Calibrated Explanations}\label{CE}
\noindent This section introduces a new feature importance explanation method called Calibrated Explanations (CE). \rev{} At the heart of the proposed method lies the realization that the feature weights must have an intuitive meaning to facilitate understanding. A \textit{factual explanation} in CE is composed of a \textit{calibrated prediction} from the underlying model accompanied by an \textit{uncertainty interval} and a collection of \textit{factual feature rules}, each composed of a \textit{feature weight with an uncertainty interval} and a \textit{factual condition}, covering that feature's instance value. \textit{Counterfactual explanations} only contain a collection of \textit{counterfactual feature rules}, each composed of a \textit{prediction estimate with an uncertainty interval} and a \textit{counterfactual condition}, covering alternative instance values for the feature. The prediction estimate represents a probability estimate for the positive class.

The following is a high-level description of how CE works: Let us assume that a scoring classifier, trained using the proper training set $Z_q$, exists for which a local explanation for test object $x$ is wanted. Let VA calibrate the underlying model for $x$ to get the probability interval $[p_0, p_1]$ and the calibrated probability estimate $p$. Use VA to estimate probability intervals \rev{($[p'_{0.f}, p'_{1.f}]$)} and calibrated probability estimates \rev{($p'_f$)} for slightly perturbed versions of object $x$, changing one feature at a time in a systematic way (see the detailed description below). To get the feature weight (and \rev{uncertainty} interval) for feature $f$, calculate the difference between $p$ to the average of all $p'_{f}$ (and $[p'_{0.f}, p'_{1.f}]$)\footnote{Exclude $p$ (and $[p_0,p_1]$), i.e., the VA results on $x$.}. 
        \begin{equation}
            w_f = p - \frac{1}{|V_f|-1}\sum p'_{f},
            \label{eq:w_f}
        \end{equation}
        \begin{equation}
            w_0^f = p - \frac{1}{|V_f|-1}\sum p'_{0.f}, 
            \label{eq:w_0^f}
        \end{equation}
        \begin{equation}
            w_1^f = p - \frac{1}{|V_f|-1}\sum p'_{1.f}.
            \label{eq:w_1^f}
        \end{equation}
\rev{The feature weight is exactly defined to be the difference between the calibrated probability estimate on the original test object $x$ and the estimated (average) calibrated probability estimate achieved on the perturbed versions of $x$. The upper and lower bounds are defined analogously using the probability intervals from the perturbed versions of $x$. As long as the same test object, underlying model and calibration set is used, the resulting explanation will also be the same.}

More formally, the following steps are pursued to achieve a factual explanation for a test object $x$:

\begin{enumerate}
  \setlength{\itemsep}{1pt}
  \setlength{\parskip}{0pt}
  \setlength{\parsep}{0pt}
    \item Use VA to get the probability interval $[p_0, p_1]$ and the calibrated probability estimate $p$ for $x$. 
    \item Separate all features into categorical features $C$ and numerical features $N$. Define a discretizer for numerical features that define thresholds and smaller- or greater-than-conditions $(\leq, >)$ for these features\footnote{\rev{This is done using a binary subclass of the \texttt{EntropyDiscretizer} class in LIME.}}. 
    \item \revised{For each feature} $f\in C$: \label{step3}
        \begin{enumerate}
              \setlength{\itemsep}{1pt}
              \setlength{\parskip}{0pt}
              \setlength{\parsep}{0pt}
            \item Iterate over all possible categorical values $v\in V_f$ and create a perturbed instance exchanging the feature value of \revised{$x_f$} with one value at a time, creating a perturbed instance \revised{$x'_{f}=v$}. 
            \item Calculate and store the probability intervals \revised{$[p'_{0.f}, p'_{1.f}]$} and the calibrated probability estimate \revised{$p'_{f}$} for the perturbed instance. \rev{Calculate the weights using equations~(\ref{eq:w_f}), (\ref{eq:w_0^f}), and (\ref{eq:w_1^f}). }
            \item \revised{Define a factual condition using the the feature $f$, the value $v$ and the identity condition $(=)$.} \label{cat_cond}
        \end{enumerate}    
    \item \revised{For each feature} $f\in N$: \label{step4}
        \begin{enumerate}
              \setlength{\itemsep}{1pt}
              \setlength{\parskip}{0pt}
              \setlength{\parsep}{0pt}
            \item Use the thresholds of the discretizer to identify the closest lower or upper threshold \revised{$t$} surrounding the feature value of \revised{$x_f$}. Divide all possible feature values in the calibration set for feature $f$ into two groups $V_f$ separated by \revised{$t$}. \label{find_threshold}
            \item Within each group, percentile values $pv$ representing the $25^{th}$, $50^{th}$ and $75^{th}$ percentiles are extracted. 
                Iterate over the values in $pv$ and create a perturbed instance exchanging the feature value of $x_f$ with one value at a time, creating a perturbed instance $x'_{f_{pv}}$. \rev{Calculate} and store the probability intervals $[p'_{0.f_{pv}}, p'_{1.f_{pv}}]$ and the calibrated probability estimate $p'_{f_{pv}}$ for the perturbed instance. 
                Average over all \rev{perturbed instances} within the group, creating a probability interval $[p'_{0.f}, p'_{1.f}]$ and the calibrated probability estimate $p'_{f}$ for each group. \rev{Calculate the weights using equations~(\ref{eq:w_f}), (\ref{eq:w_0^f}), and (\ref{eq:w_1^f}). }
            \item \revised{Define a factual condition using threshold $t$ and feature $f$. The $\leq \text{or} >$ condition is used so that the factual condition covers the instance value.} \label{num_cond}
        \end{enumerate} 
\end{enumerate}   
        

It is essential to discuss \rev{rule logic that results in unambiguously interpretable rules. For categorical features, the only logical factual rule is based on the identity condition. For numerical features, several different rules can be formed.} As the feature weights are the difference between the calibrated probability estimate \rev{of $x$} and the average \rev{calibrated probability estimates on the perturbed instances within} a group, it would obscure the meaning of the feature weight if the feature rule could \rev{be a \textit{between} rule, i.e.,} be of the form $0<$~\verb|feature f|~$\leq2$. The reason is that it is often reasonable to assume that probabilities for values below the interval (\verb|feature f|~$\leq0$) may differ significantly from probabilities for values above the interval (\verb|feature f|~$>2$), making an average of the two hard to interpret. Consequently, \rev{it is only the threshold being closest to the feature value that is used to form a rule.}

\rev{The worst case computational complexity of the method is determined by the following things:
\begin{itemize}
              \setlength{\itemsep}{1pt}
              \setlength{\parskip}{0pt}
              \setlength{\parsep}{0pt}
    \item Training of the underlying model and calculating the probability estimates on the calibration set. This is done once and the complexity is determined by the underlying model. 
    \item Initializing VA once, which involves sorting and pre-computing some values, taking at worst time $O(l\text{ log }l$), i.e., being dependent on the size $l$ of the calibration set \citep{petej2018thesis}. 
    \item The method operates feature-by-feature and iterates over all categorical values for categorical features or over the number of percentile samples for numerical features. Each iteration makes a call to VA, which has a complexity of $O(\text{log }l)$ for producing the calibrated probability interval \citep{petej2018thesis}. Thus, the complexity for the method is $O(|F|i\text{ log }l)$, where $|F|$ is the number of features and $i$ is the number of iterations (i.e., the number of values or samples) per feature. Normally, there are a lot fewer features than instances, so that $|F|<<l$. The number of values or samples are also generally small (the default number of samples for numerical features is 3).
    \item In connection to the call to VA, the underlying model is also called once to get the probability estimate. This is determined by the underlying model.
\end{itemize}}

By definition, CE calibrates the underlying model using VA, creating well-calibrated predictions \rev{with uncertainty quantification. The way that CE utilizes VA enables the method to also provide} uncertainty quantification of feature weights. By using equality rules for categorical features and binary rules for numerical features, interpreting the feature weights in relation to the rules and the instance values is straightforward and unambiguous. As opposed to LIME and SHAP, which rely on the creation of large sets of perturbed instances, CE instead rely on a calibration set which must be taken from the same distribution as the test instances. CE is relying on insights gained from the calibration set about the model when calibrating it's probability estimates. The reliance on a calibration set from the same distribution as the test object will contribute to robust explanations as they will always be smoothed by the underlying distribution shared between test and calibration data. Depending on the size of the calibration set, which is used to train a few isotonic regression models per feature, the generation of well-calibrated explanations is, in most cases, fast compared to existing solutions such as LIME and SHAP.

\subsection{Counterfactual Calibrated Explanations}
\noindent Using the definition of CE above, it becomes straightforward to generate counterfactual rules. The main difference when using Counterfactual Calibrated Explanations (CCE) is that the conditions defined in steps~\ref{cat_cond} and~\ref{num_cond} are counterfactual rather than factual. This means that the categorical counterfactual condition (step~\ref{cat_cond}) is using the $\neq$ condition. Numerical counterfactual conditions (step~\ref{num_cond}) are defined as $\leq$-conditions and $>$-conditions but with conditions that exclude the instance value. Since non-binary discretizers are \revised{used} for numerical features\revised{, numerical counterfactual conditions will} often allow both \revised{counterfactual $\leq$-conditions and $>$-conditions} to be formed\footnote{\rev{All existing Discretizers in LIME can be used for counterfactual explanations. The \texttt{EntropyDiscretizer} class is used by default.}}. Probability intervals \revised{$[p'_{0.f}, p'_{1.f}]$} are defined following the CE procedure described in \revised{steps~\ref{step3} and~\ref{step4} above. It} is trivial to define a \revised{counterfactual} rule for each of the alternative values, with the expected probability interval for each rule already defined \revised{by $[p'_{0.f},p'_{1.f}]$}. The feature weights defined in equation~(\ref{eq:w_f}) are only used for sorting counterfactual rules based on impact\revised{, i.e., the distance to $p$}. 

\rev{In classification, counterfactual explanations normally refer to a method that provide the minimal set of changes necessary to alter the predicted class. CE and CCE are providing explanations in terms of probability estimates, making a definition of counterfactual explanations for continuous targets more relevant. Consequently,} CCE is a counterfactual explanation in the sense that it creates rules that describe the alterations stemming from counterfactual conditions. Since the explanations operate on probability estimates rather than class labels, any adjustment to a probability estimate (or the probability intervals) can be considered an alternative outcome, even when it does not alter the predicted class label. It is important to note that while CCE showcases alterations in feature values critical to the prediction, it may not explicitly identify the true counterfactual for numerical features (i.e., the minimal value resulting in a change). CCE emphasizes the most impactful changes in driving the model's prediction, providing a nuanced perspective on the model's decision-making process.

\subsection{The \texttt{calibrated-explanations} Package}
\noindent The proposed method \href{https://github.com/Moffran/calibrated_explanations}{\texttt{calibrated-explanations}}\footnote{\rev{\href{https://github.com/Moffran/calibrated_explanations}{https://github.com/Moffran/calibrated\_explanations}}} is published as an open source Python package. The version described here is \textit{v0.2.3}. The package can be installed in the following ways: 
\begin{itemize}
              \setlength{\itemsep}{1pt}
              \setlength{\parskip}{0pt}
              \setlength{\parsep}{0pt}
    \item \texttt{pip install calibrated-explanations}
    \item \texttt{conda install -c conda-forge calibrated-explanations}
\end{itemize}

To explain an already trained model, a \texttt{CalibratedExplainer}, taking the model and a calibration set as input, can be used. To create a collection of calibrated explanations, the \texttt{explain\_factual()} or \texttt{explain\_counterfactual()} methods, taking the test objects to be explained as input, is used. As an alternative, a \texttt{WrapCalibratedExplainer}, taking an untrained model, can be used to train (using the \texttt{fit(train\_X, train\_y)} function), calibrate and create explanations in an integrated way.

The \texttt{explain\_factual()} or \texttt{explain\_counterfactual()} methods output a collection of explanations, \texttt{CalibratedExplanations}, containing a set of \texttt{FactualExplanation} or \texttt{CounterfactualExplanation} objects. The whole collection of individual explanations can be plotted using the \texttt{plot\_all()} method. To access individual explanations, \texttt{get\_explanation(index)} can be used on the collection. \texttt{plot\_explanation()} can be used on individual explanations and \texttt{plot\_explanation(index)} can be used on the collection. When plotting a \texttt{FactualExplanation}, all plotting functions accept an additional parameter \texttt{uncertainty=True} to create uncertainty plots.

For further details regarding how to use the package, see the \href{https://calibrated-explanations.readthedocs.io/en/latest/?badge=latest}{documentation} or the \href{https://github.com/Moffran/calibrated_explanations/tree/main/notebooks}{notebook folder} for examples. The Venn-Abers implementation used is the \href{https://github.com/ip200/venn-abers}{\texttt{venn-abers}} package by Ivan Petej \citep{vovk2015largescale,vovk2014vennabers}.

\section{Presentation of Calibrated Explanations}\label{presentation}
\noindent CE \revised{and CCE }can be analysed from different perspectives. In this paper, three different kinds of plots are \revised{used to illustrate factual and counterfactual explanations}. The first two are used \revised{for factual} explanations:
\begin{itemize}
              \setlength{\itemsep}{1pt}
              \setlength{\parskip}{0pt}
              \setlength{\parsep}{0pt}
    \item Regular explanations, providing well-calibrated explanations without any uncertainty information. These explanations are directly comparable to other feature importance explanation techniques, like LIME. 
    \item Uncertainty explanations, providing well-calibrated explanations including uncertainty intervals to highlight both the importance of a feature and the amount of uncertainty connected with its estimated importance. \revised{Uncertainty is visualized by adding the parameter \texttt{uncertainty=True} to the plot functions.}
\end{itemize} 

These plots are inspired by LIME. Especially the rules in LIME have been seen as valuable information in the explanations. For the reasons given above, CE \revised{create} binary rules with \revised{factual} explanations. 

The third kind of plot is a counterfactual plot showing how the probability estimate \revised{is affected} when other feature values are used.

When plotting \revised{CE and} CCE explanations, the user can choose to limit the number of rules to show\revised{. For factual explanations, the number of rules} equals the number of features, as there is one rule per feature. However, in counterfactual explanations, where CCE creates as many counterfactual rules as possible, rules are ordered based on impact, starting with the most impactful rules.

\subsection{Factual Calibrated Explanations}
\subsubsection{Without uncertainty}
\noindent Regular CE explanations are similar to LIME explanations in several ways. First, every feature has a corresponding conditional rule that identifies the condition under which the weight is defined \rev{(written out at the left side of the plot)}. Secondly, every rule has a corresponding feature weight, which is defined in such a way that the weight \rev{plus} the probability achieved if the rule condition is broken results in the probability of the instance covered by the rule. As a consequence, the meaning of the rule and the weight it produces can be unambiguously understood. 

\begin{figure}[htbp!]
  \centering
    \includegraphics[width=\textwidth]{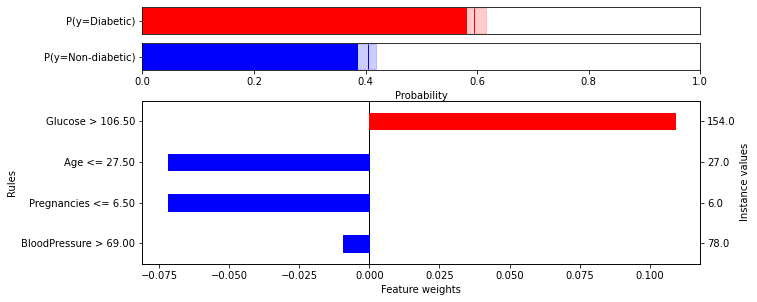}
    \caption{Regular CE plot with continuous feature values from the data set Diabetes. \revised{The factual conditions are shown to the left and the actual feature values are shown to the right of the feature plot.}}
    \label{fig:CE_reg_mean_Diab}
\end{figure}
Figure~\ref{fig:CE_reg_mean_Diab} shows a regular CE plot from the diabetes data set with a calibrated probability estimate of about $0.6$ for the positive class, indicating that the patient should be classified as \textit{diabetic} (shown by the two bars at the top). The main plot visualizes the feature weights, \revised{centered at} the solid black line at $0$\revised{, which corresponds to} the calibrated probability estimate \revised{$p$ for the positive class}\rev{, indicated by the solid line in the \textit{P(y=Diabetic)} bar}. Among the different features, \rev{one feature clearly contribute positively towards the patient being \textit{diabetic} whereas the remaining three features contribute towards the patient being \textit{non-diabetic}. In the given example, the feature \textit{Glucose} influence the prediction the most, with a weight of $0.11$ in favour of \textit{diabetic}, as long as the feature value stays above $106.5$. \textit{Age} and \textit{Pregnancies} are the second and third most influential features, where the low age and the number of pregnancies being less than $6.5$ both influence the prediction towards \textit{non-diabetic} with about $0.07$. \textit{BloodPressure} is the fourth feature that is identified as having an impact on the prediction, also favoring \textit{non-diabetic}. The ambiguous prediction, with fairly eevn probability estimates, can be understood from the fact that the weights point in opposite directions.}

\begin{figure}[h!]
    \centering
    \includegraphics[width=\textwidth]{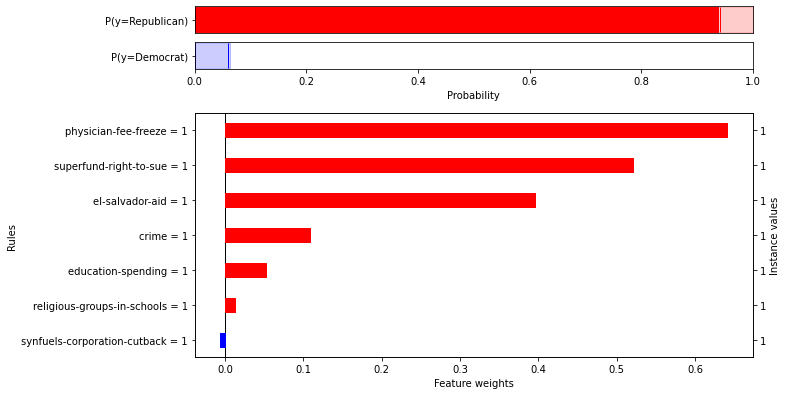}
    \caption{Regular CE plot with categorical feature values from the data set Vote.}
    \label{fig:CE_reg_mean_Vote}
\end{figure}

With a data set where the features have got categorical values and a binary target \rev{(\textit{Democrat} or \textit{Republican})}, as in figure~\ref{fig:CE_reg_mean_Vote}, the rules look somewhat different. In the given example, the rules are either equal to $0$ \revised{or} $1$, with the feature values being covered by the rule. The data set is taken from results of the voting in \revised{US} Congress 1984, where $0$ indicated the congressman had voted negatively to a law proposal and a $1$ that the vote was positive. In the figure, the top feature \textit{physician-fee-freeze} is most influential in explaining why this congressman is predicted as \textit{Republican}. \rev{Almost all features favor the positive class, explaining why this prediction has a high probability estimate for the positive class.}

\subsubsection{With uncertainty}
\begin{figure}[htbp!]
    \centering
    \includegraphics[width=\textwidth,trim={0 0 0 0},clip]{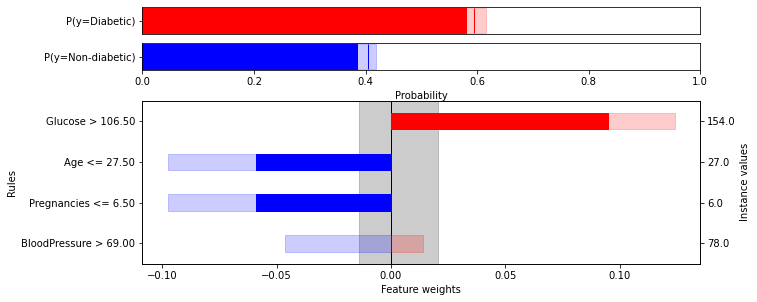}
    \caption{Uncertainty CE plot with intervals from the data set Diabetes.}
    \label{fig:Ce_intervals_diab}
\end{figure}

\noindent \revised{Since CE creates an uncertainty interval for the weights, it is possible to learn from the uncertainty. When uncertainty is shown by setting the parameter \texttt{uncertainty=True},} CE adds intervals to the plot, showing the uncertainty of the prediction\revised{. A} narrow interval indicates certainty in the estimate or weight, and a wider interval is more uncertain in the estimate or weight. Moreover, the feature interval also shows how the feature affects the prediction estimate. \revised{F}igure~\ref{fig:Ce_intervals_diab} presents the \rev{explanation of the same instance} as in figure~\ref{fig:CE_reg_mean_Diab} \revised{with uncertainty added}. 

The \rev{probability} interval for the underlying model is indicated both in the probability bars at the top, as light blue or light red, and as the light grey area in the main part of the plot. Similarly, the feature weights are solidly coloured, from the black line, representing the calibrated probability estimate, to the lower bound of the weight interval and lightly coloured between the lower and upper bound of the interval. The rules are ordered based on the feature weights of the calibrated prediction. Blue bars indicate that the feature support \rev{the \textit{non-diabetic}} class, and the red bars indicate support for \rev{the \textit{diabetic} class}. \rev{The amount of uncertainty of the different feature weights vary slightly, with the least important feature having the most uncertainty. The uncertainty interval for \textit{BloodPressure} even covers the original probability estimate, indicating an uncertainty regarding if the feature affects the prediction at all.}

\begin{figure}[htbp!]
    \centering
    \includegraphics[width=\textwidth]{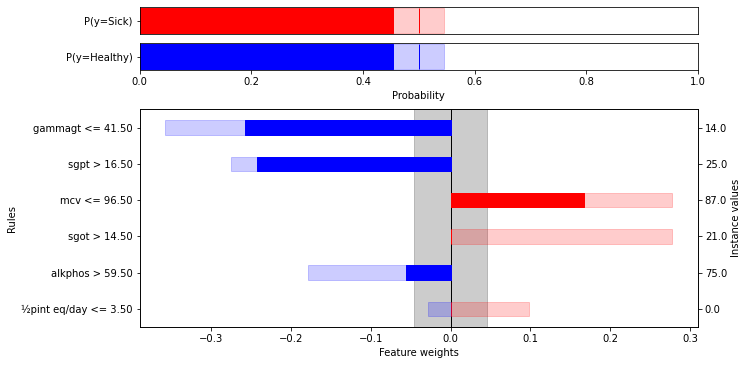}
    \caption{Uncertainty CE plot with intervals from the data set Liver.}
    \label{fig:Ce_intervals_liv}
\end{figure}

A different situation is seen in figure~\ref{fig:Ce_intervals_liv} where the probability of not having a liver disease is pending between the two classes. \rev{The weights point in both directions and the uncertainty intervals vary a lot. The upper bound for the \textit{mcv} and \textit{sgot} features are the same while the lower bound for \textit{sgot} is much lower, indicating a much larger degree of uncertainty.}

\subsection{Counterfactual Calibrated Explanations}
\noindent In CCE, the plots are not showing feature weights. Instead, the plot is focusing on the VA probability intervals. Each rule shows the alternative VA probability interval resulting from \rev{perturbing} the feature value to a value covered by the counterfactual rule condition. 
\begin{figure}[h!]
    \centering
    \includegraphics[width=\textwidth]{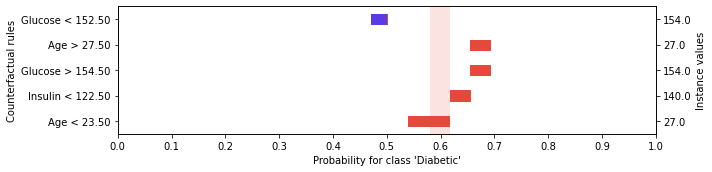}
    \caption{CCE counterfactual plot with continuous feature values from the data set Diabetes (the same instance as in figure~\ref{fig:Ce_intervals_diab}). \revised{The counterfactual conditions are shown to the left and the feature values are shown to the right.}}
    \label{fig:CE_counterfactual_diab}
\end{figure}
Numerical features can, at most, result in two counterfactual rules (above or below the thresholds surrounding the feature value), whereas one counterfactual rule is created for each alternative categorical value. In the plots, only the \rev{counterfactual rules affecting the probability estimate} are shown. 

Like the uncertainty plot, the counterfactual plot shows that the two features \rev{\textit{Glucose} and \textit{Age} affect the prediction most. However, in the plot, the rules show that even a small reduction in the \textit{Glucose} feature (\texttt{Glucose}~$<152.5$) may change the probability estimate enough to change the prediction of the model. Higher \textit{Age} and \textit{Glucose} will increase the likelyhood for the patient being \textit{diabetic}.}

\begin{figure}[h!]
    \centering
    \includegraphics[width=\textwidth]{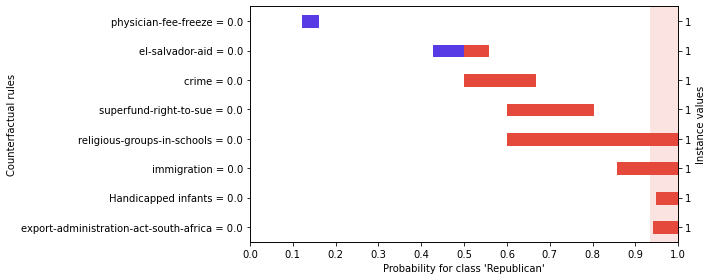}
    \caption{CE counterfactual plot with categorical feature values from the data set Vote.}
    \label{fig:CE_counterfactual_vote}
\end{figure}

In figure~\ref{fig:CE_counterfactual_vote}, the chosen instance is the same as in figure~\ref{fig:CE_reg_mean_Vote} with a clear prediction of the congressman being a \textit{Republican}. There are some interesting features to point out for this instance. First, just as seen earlier in figure~\ref{fig:CE_reg_mean_Vote}, the \rev{\textit{physician-fee-freeze} feature is highly important for the prediction and changing the value according to the counterfactual rule would modify the prediction. The second counterfactual rule would decrease the certainty dramatically without clearly indicating whether the model would predict it as \textit{Democrat} or \textit{Republican}. Several of the remaining counterfactual rules also affect the probability and uncertainty a lot but without changing the prediction.}


\section{Experiments}\label{evaluation}
\subsection{Experimental Setup}
\noindent \rev{In order to verify some of the claims of CE, two experiments have been run. The two experiments evaluate the stability and the robustness of the method. Stability is evaluated through experiments where the same model, calibration set and test set have been explained 30 times per data set. The only source of variation was the random seed. Robustness is evaluated through experiments where training and calibration sets have been randomly resampled before a new model was trained and explained. The experiment where run 30 times per data set with the same test set for all runs. A test set with 20 stratified instances (making sure both classes being equally represented) were used. Robustness is measured in this way to avoid inferring perturbed instances which are not from the same distribution as the test instances being explained. The probability estimate of each of the models was computed on the same test set, as comparison to the robustness results. The expectation is that a stable and robust explanation method should result in low variance in the feature weights. 
Both random forests and xGBoost are used and both factual and counterfactual explanations are evaluated. 
}


$25$ binary classification problems were used, publicly available from either the UCI repository \citep{uci} or the PROMISE Software Engineering Repository \citep{promise}. The data set's characteristics are presented in Table~\ref{tab:datasets}, where \textit{\#inst.} is the number of instances and \textit{\#attrib.} is the number of input attributes. \rev{Attributes with less than $10$ unique values were treated as categorical attributes.}
\setlength{\tabcolsep}{2pt}
\begin{table}[H]
\caption{Data set characteristics}
\label{tab:datasets}  
\begin{center} 
\footnotesize
\noindent
\begin{tabular}{lcc|lcc}
Name & \#inst. & \#attrib.  & Name & \#inst. & \#attrib. \tabularnewline
\hline 
colic    & 357  & 60    & kc2           & 369   & 22 \tabularnewline
creditA  & 690  & 43    & kc3           & 325   & 40 \tabularnewline
diabetes & 768  & 9     & liver         & 341   & 7  \tabularnewline
german   & 955  & 28    & pc1req        & 104   & 9  \tabularnewline
haberman & 283  & 4     & pc4           & 1343  & 38 \tabularnewline
heartC   & 302  & 23    & sonar         & 208   & 61 \tabularnewline
heartH   & 293  & 21    & spect         & 218   & 23 \tabularnewline
Hearts   & 270  & 14    & spectf        & 267   & 45 \tabularnewline
hepati   & 155  & 20    & transfusion   & 502   & 5  \tabularnewline
iono     & 350  & 34    & tictactoe     & 958   & 28 \tabularnewline
je4042   & 270  & 9     & vote          & 517   & 17 \tabularnewline
je4243   & 363  & 9     & wbc           & 463   & 10 \tabularnewline 
kc1      & 1192 & 22 \tabularnewline
\end{tabular}
\end{center}
\indent
\normalsize
\end{table}

\rev{The code used to run the experiments are located in the \href{https://github.com/Moffran/calibrated_explanations/tree/main/evaluation}{evaluation} folder in the repository together with supplementary material. The experiment files for this paper are named \texttt{Classification\_[..]\_stab\_rob}.}

\subsection{Results}
\noindent \rev{The results in Table~\ref{tab:stab_rob} are the mean variance of the stability and robustness measured over the 30 runs and 20 instances. The variance is measured per instance and computed over the 30 runs on the feature weight of the most influential feature, defined as the feature most often having highest absolute feature weight. The average variance is computed over the 20 instances. The most influential feature is used since it is the feature that is most likely to be used in a decision but also the feature with the greatest expected variation (as a consequence of the weights having the highest absolute values). The \textit{Model} column is the mean variance of the underlying model's probability estimate for the robustness experiment prior to calibration. }
\begin{table}[H]
    \centering
    \footnotesize
    \begin{tabular}{l|cccc|cccccc}
         & \multicolumn{4}{c}{Stability} & \multicolumn{6}{|c}{Robustness} \\
         & \multicolumn{2}{c}{xGB} & \multicolumn{2}{c}{RF} & \multicolumn{3}{|c}{xGB} & \multicolumn{3}{c}{RF} \\
        Dataset & CE & CCE & CE & CCE & CE & CCE & Model & CE & CCE & Model \\
\hline
colic & 3.9e-35 & 3.9e-35 & 7.8e-33 & 7.8e-33 & 0.015 & 0.015 & 0.015 & 0.017 & 0.017 & 0.010 \\
creditA & 1.5e-34 & 1.5e-34 & 1.5e-32 & 1.5e-32 & 0.022 & 0.022 & 0.011 & 0.015 & 0.015 & 0.009 \\
diabetes & 1.2e-33 & 1.2e-33 & 5.0e-33 & 5.0e-33 & 0.017 & 0.017 & 0.014 & 0.015 & 0.015 & 0.013 \\
german & 1.2e-34 & 1.2e-34 & 9.6e-34 & 9.6e-34 & 0.003 & 0.003 & 0.015 & 0.005 & 0.005 & 0.014 \\
haberman & 5.2e-34 & 5.2e-34 & 1.2e-33 & 1.2e-33 & 0.010 & 0.010 & 0.016 & 0.011 & 0.011 & 0.016 \\
heartC & 0.0e+00 & 0.0e+00 & 2.7e-33 & 2.7e-33 & 0.012 & 0.012 & 0.017 & 0.012 & 0.012 & 0.016 \\
heartH & 2.6e-33 & 2.6e-33 & 4.9e-33 & 4.9e-33 & 0.017 & 0.017 & 0.017 & 0.011 & 0.011 & 0.016 \\
heartS & 1.6e-33 & 1.6e-33 & 4.3e-33 & 4.3e-33 & 0.019 & 0.019 & 0.018 & 0.017 & 0.017 & 0.017 \\
hepati & 1.9e-34 & 1.9e-34 & 5.2e-33 & 5.2e-33 & 0.022 & 0.022 & 0.018 & 0.014 & 0.014 & 0.017 \\
iono & 4.1e-35 & 4.1e-35 & 3.6e-33 & 3.6e-33 & 0.028 & 0.028 & 0.017 & 0.013 & 0.013 & 0.016 \\
je4042 & 1.5e-33 & 1.5e-33 & 6.4e-33 & 6.4e-33 & 0.018 & 0.018 & 0.017 & 0.015 & 0.015 & 0.017 \\
je4243 & 5.3e-34 & 5.3e-34 & 5.0e-33 & 5.0e-33 & 0.010 & 0.010 & 0.018 & 0.010 & 0.010 & 0.017 \\
kc1 & 5.4e-34 & 5.4e-34 & 1.7e-33 & 1.7e-33 & 0.011 & 0.011 & 0.017 & 0.009 & 0.009 & 0.017 \\
kc2 & 7.7e-34 & 7.7e-34 & 5.0e-34 & 5.0e-34 & 0.018 & 0.018 & 0.018 & 0.007 & 0.007 & 0.017 \\
kc3 & 1.4e-35 & 1.4e-35 & 5.7e-34 & 5.7e-34 & 0.007 & 0.007 & 0.017 & 0.005 & 0.005 & 0.017 \\
liver & 2.2e-33 & 2.2e-33 & 8.7e-33 & 8.7e-33 & 0.025 & 0.025 & 0.017 & 0.030 & 0.030 & 0.017 \\
pc1req & 7.7e-35 & 7.7e-35 & 1.8e-33 & 1.8e-33 & 0.014 & 0.014 & 0.018 & 0.014 & 0.014 & 0.017 \\
pc4 & 3.7e-34 & 3.7e-34 & 2.5e-33 & 2.5e-33 & 0.009 & 0.009 & 0.017 & 0.007 & 0.007 & 0.017 \\
sonar & 1.8e-34 & 1.8e-34 & 2.2e-33 & 2.2e-33 & 0.019 & 0.019 & 0.017 & 0.009 & 0.009 & 0.017 \\
spect & 0.0e+00 & 0.0e+00 & 1.5e-33 & 1.5e-33 & 0.007 & 0.007 & 0.017 & 0.006 & 0.006 & 0.016 \\
spectf & 4.6e-34 & 4.6e-34 & 4.2e-34 & 4.2e-34 & 0.006 & 0.006 & 0.016 & 0.005 & 0.005 & 0.016 \\
transfusion & 2.8e-34 & 2.8e-34 & 2.9e-33 & 2.9e-33 & 0.009 & 0.009 & 0.017 & 0.008 & 0.008 & 0.017 \\
ttt & 0.0e+00 & 0.0e+00 & 4.2e-32 & 4.2e-32 & 0.017 & 0.017 & 0.016 & 0.025 & 0.025 & 0.016 \\
vote & 0.0e+00 & 0.0e+00 & 6.5e-33 & 6.5e-33 & 0.014 & 0.014 & 0.016 & 0.011 & 0.011 & 0.016 \\
wbc & 5.7e-33 & 5.7e-33 & 2.5e-33 & 2.5e-33 & 0.021 & 0.021 & 0.016 & 0.018 & 0.018 & 0.015 \\
\hline
Average & 7.6e-34 & 7.6e-34 & 5.4e-33 & 5.4e-33 & 0.015 & 0.015 & 0.016 & 0.012 & 0.012 & 0.016 \\
    \end{tabular}
    \caption{Stability and Robustness}
    \label{tab:stab_rob}
\end{table}
\rev{As can be seen Table~\ref{tab:stab_rob}, the stability is practically $0$ for both factual CE (CE) and counterfactual CE (CCE), illustrating that the method is stable by definition. The minuscule variations stem from the different random seeds, having a negligible impact. }

\rev{The robustness is also low, even if the mean variance robustness is clearly larger than for stability. The robustness is comparable to the variance of the probability estimates used as reference. This indicates that the method is fairly robust to perturbations such as variations of the calibration set and the model. Obviously, since the method is explaining the calibrated probability estimates of the model, it must be expected that the method is sensitive to changes to the probability estimates. } 

\section{Discussion}
\subsection{Characteristics of Calibrated Explanations}
\noindent \rev{One of the most important characteristics of a good explanation method is that it must be \textit{reliable}. When it comes to XAI methods, reliability most often refers to its fidelity to the model being explained. However, if the model in itself is not reliable, i.e., if it is poorly calibrated and thus misrepresents reality, high fidelity is more likely to convey harm than benefit. Consequently, to create reliable explanations, the explanation must have high fidelity to a model which is well-calibrated. In addition, there is yet another layer to reliability which relates to uncertainty. For an explanation to be truly reliable, the explanation must be able to convey insights into when the model is making a prediction in which it has high confidence contra when it has low confidence. This goes beyond the model having high or low probability estimates, since the probability estimate may be high, but the model may not have seen many similar cases and thus have low confidence in the prediction, and vice versa. The main strength of the proposed method is that it provides high reliability with regard to all three layers of reliability.
\begin{itemize}
              \setlength{\itemsep}{1pt}
              \setlength{\parskip}{0pt}
              \setlength{\parsep}{0pt}
    \item The model is calibrated to ensure an accurate reflection of reality. This includes uncertainty quantification of the model's probability estimates.
    \item The explanations have high fidelity with feature weights exactly defined in relation to the calibrated probability estimates of the model. The feature weights are also calibrated to ensure fidelity.
    \item The method produces uncertainty quantification of the feature weights that provide insights about the degree of confidence to put in each weight. For the counterfactual explanations, the method instead provide uncertainty quantification about the estimated probability interval.
\end{itemize}}

\rev{\textit{Stability} is, as mentioned earlier, a critical aspect of an explanation method and something that primarily LIME has been criticized for lacking (see, e.g., \cite{rahnama2019study}). The method is stable by design, which is clearly shown in the experiments. The explanation generated for an instance using CE or CCE is stable as long as the same calibration set and model are used. }

\rev{Another important characteristic is \textit{robustness} to change. The explanations are robust by design, i.e., consistent, as long as the feature rules cover perturbations in the input values and the model prediction is not affected more than marginally. Obviously, if the model prediction is affected a lot by a perturbation, it cannot be considered a small perturbation in this context, as it has a great impact on the model and thus should also have a great impact on the explanation for the method to be reliable. The reliance on the calibration set from the same distribution as the test object will contribute to robust explanations as they will always be smoothed by the underlying distribution shared between test and calibration data. The experiment, evaluating how the method is affected by changes to calibration and model predictions, shows that the method updates its feature weights in accordance with how much the underlying model is changing its predictions.} 

\subsection{\rev{Calibrated Explanations vs. SOTA methods}}
\noindent \rev{The method presented in this paper generates well-calibrated local explanations with uncertainty quantification of both the probability estimate of the model and of the feature weights. As seen in Table~\ref{tab:SOTA_vs_CE}, it is only BayesSHAP and BayesLIME that share the possibilities of getting calibrated explanations and uncertainty quantification. These methods are instantiations of a Bayesian framework into SHAP and LIME rather than new methods. }

\rev{None of the SOTA methods could generate both factual and counterfactual explanations (see Table~\ref{tab:SOTA_vs_CE}), further highlighting the all-inclusive nature of CE. It is worth pointing out that none of the counterfactual methods were generating neither well-calibrated explanations nor uncertainty quantification.}

\begin{table}[t!]
  \centering
  \footnotesize
  \caption{Characteristics of CE vs factual and counterfactual SOTA Explanation Methods \citep{Ribeiro2016_kdd,lundberg2017unified,plumb2018model, ribeiro2018anchors,slack2021reliable,guidotti2018local,karimi2020model}.}
  \begin{tabular}{l|ccccccc}
    & & & & \textbf{Cnter} & & \textbf{Model}& \textbf{Uncert.} \\
    \textbf{Method} & \textbf{Local} & \textbf{Global} & \textbf{Factual} & \textbf{factual}& \textbf{Calibr.}& \textbf{agnost.}& \textbf{quant.} \\
    \hline
    CE &    X & - & X & X & X & X & X\\
    LIME &  X & - & X & - & - & X & - \\
    SHAP & X & X & X & - & - & X & -\\
    MAPLE & X & - & X & - & - & X& -\\
    BayesLIME &  X & - & X & - & X & X & X \\
    BayesSHAP & X & X & X & - & X & X & X\\
    ANCHOR & X & - & X & - & - & X & - \\
    MACE & X & - & - & X & - & X & - \\
    LORE & X & - & - & X & - & X & - \\
    \end{tabular}%
  \label{tab:SOTA_vs_CE}%
\end{table}%

\section{Conclusion}

\noindent This paper introduces \textit{Calibrated Explanations} (CE), a novel explanation method that not only calibrates the underlying model but also provides uncertainty information in the explanations. Each feature's weight is defined by its contribution to the calibrated probability estimate for the positive class, and the explanations are conveyed through straightforward conditional rules.

CE does not stop at factual explanations; it inherently enables the creation of counterfactual rules, offering insights into the uncertainty of each counterfactual outcome. The method is available as an open-source package, installable via \texttt{pip} or \texttt{conda}, supporting the plotting of both factual and counterfactual explanations.

\rev{Two experiments evaluating the method's robustness and stability were conducted on 25 benchmarking datasets. In the evaluations, CE} demonstrated outstanding performance: effective in terms of running time, reliable due to inherent calibration and uncertainty quantification, stable with consistent calibration sets and models, and robust against small perturbations. In essence, Calibrated Explanations embody the hallmarks of a high-quality explanation method.

\rev{CE generates both factual and counterfactual explanations, which is rather unique compared to other SOTA methods. None of the methods studied in the paper could offer such a wide range of characteristics like CE: either they focus on factual or counterfactual explanations. Apart from CE, only one approach (with two instantiations) included uncertainty information and calibrated explanations.}

Looking ahead, future work includes real-world scenario evaluations for user perception and support for image and text data. The evolution of the currently available plots and the introduction of additional visualization methods to enhance rule insights are also key priorities.

\section*{Acknowledgements}
\noindent Helena Löfström is a PhD student in the Industrial Graduate School in Digital Retailing (INSiDR) at the University of Borås, funded by the Swedish Knowledge Foundation, grant no. 20160035. The authors acknowledge the Swedish Knowledge Foundation and the industrial partners for financially supporting the research and education environment on Knowledge Intensive Product Realization SPARK at Jönköping University, Sweden. Projects: AFAIR grant no. 20200223 and PREMACOP grant no. 20220187. 

\section*{Competing interests}
\noindent The authors acknowledge that we have no competing interests to declare.

\section*{Declaration of Generative AI and AI-assisted technologies in the writing process}
\noindent During the preparation of this work the author(s) used ChatGPT, Bard and Bing Chat in order to get suggestions on how to improve and simplify descriptions and sentences. After using this tool/service, the author(s) reviewed and edited the content as needed and take(s) full responsibility for the content of the publication.

\bibliography{ARXIV}
\end{document}